\documentclass[conference]{IEEEtran}
\usepackage{cite}
\usepackage{graphicx}

\usepackage{amsmath} 
\usepackage{bm}
\usepackage[]{algorithm2e}
\usepackage{graphicx}
\usepackage[lofdepth,lotdepth]{subfig}

\usepackage{tikz}
\usetikzlibrary{shapes,arrows}

\usepackage{booktabs}
\newcommand{\ra}[1]{\renewcommand{\arraystretch}{#1}}

\newcommand{\YLD}{\mbox{YLD}}
\newcommand{\YLL}{\mbox{YLL}}
\newcommand{\DALY}{\mbox{DALY}}
\newcommand{\TTC}{\mbox{TTC}}
\newcommand{\TRC}{\mbox{TRC}}
\newcommand{\HSC}{\mbox{HSC}}

\begin{document}


\title{Novel Exploration Techniques (NETs) for Malaria Policy Interventions}

\author
{\IEEEauthorblockN{Oliver Bent}
\IEEEauthorblockA{University of Oxford\\
oetbent@robots.ox.ac.uk}
\and
\IEEEauthorblockN{Sekou L. Remy}
\IEEEauthorblockA{IBM Research Africa\\
sekou@ke.ibm.com}
\and
\IEEEauthorblockN{Stephen Roberts}
\IEEEauthorblockA{University of Oxford\\
sjrob@robots.ox.ac.uk}
\and
\IEEEauthorblockN{Aisha Walcott-Bryant}
\IEEEauthorblockA{IBM Research Africa\\
awalcott@ke.ibm.com}
}

\maketitle


\begin{abstract}
The task of decision-making under uncertainty is daunting, especially for problems which have significant complexity. 
Healthcare policy makers across the globe are facing problems under challenging constraints, with limited tools to help them make data driven decisions.
In this work we frame the process of finding an optimal \textit{malaria} policy as a stochastic multi-armed bandit problem, and implement three agent based strategies to explore the policy space.
We apply a Gaussian Process regression to the findings of each agent, both for comparison and to account for stochastic results from simulating the spread of malaria in a fixed population. 
The generated policy spaces are compared with published results to give a direct reference with human expert decisions for the same simulated population.
Our novel approach provides a powerful resource for policy makers, and a platform which can be readily extended to capture future more nuanced policy spaces.
\end{abstract}


\section{Introduction}
Malaria is a mosquito-borne disease which is endemic in sub-Saharan Africa (SSA).
There has been a significant progress in the prevention and control of the disease resulting in a reduction of the mortality rate of malaria and the number of new cases. 
Many countries in SSA rely heavily on external funding for malaria prevention and control which, in recent years, investments have started to level-off \cite{Winskill2011}. 
This means that policy makers will have more difficult decisions to ensure continued successes in the management of the disease and of their populations given the available resources.
Moreover, it is projected that up to \$450M in research and development (R\&D) is required each year for malaria up to 2018, with slower growth needed thereafter \cite{moran2007malaria}. 
It is critical for SSA countries to have access to tools to develop policies that maximize the cost-effectiveness of malaria interventions. 

Individual distributed decision makers (e.g., NGOs, governments and charities) must be able to explore the possible set of actions for appropriate malaria interventions within their populations.  
Such policies include a mix of actions like the distribution of long-lasting insecticide-treated nets (ITNs), indoor residual spraying (IRS), vector larvicide in bodies of water, and malaria vaccinations.  
The space of possible policies for malaria interventions is daunting and inefficient for human decision makers to explore without adequate decision support tools. 

In this work we describe a novel formulation for the systematic exploration of malaria intervention actions.
This formulation is well-suited for applying Artificial Intelligence (AI) agents to learn the most effective intervention strategies for a specific environment. 
To date, the applications of AI in healthcare have focused around prediction of disease spread, diagnosis, and personalized care planning tools (e.g., \cite{Piette2016} and \cite{Dimitrov2009}).
Building on these applications, our work leverages the OpenMalaria Platform \cite{SMITH2008a}, which provides a simulation environment for an agent to learn optimal policies for the control of the disease. 
The OpenMalaria codebase gives access to stochastic transmission models of malaria and can be used by researchers to evaluate the impact of various malaria control interventions. OpenMalaria therefore provides a platform to create a simulation environment from which an agent may explore optimal policies for the control of malaria transmission. Specifically, the work presented will make use of a parameterisation of OpenMalaria models which describes the Rachuonyo South district in Western Kenya \cite{Stuckey2012}.

Our approach is to apply multiple agents to determine the optimal malaria policy based on any combination of coverage of ITN and IRS for the target population. 
The reward function is determined by an application of the cost of disability adjusted life years. 
A key benefit of this work is analytical search space exploration to converge on an optimal policy. 
We demonstrate how agent-based exploration techniques and advances in compute infrastructure can be leveraged to determine the optimal policy of malaria interventions for a particular environment, without expert human guidance. 
Moreover, our work shows the potential for a systematic agent based decision support system for human decision makers exploring cost-effective intervention strategies.  

\section{Stochastic Multi-Armed Bandit}

Finding an optimal malaria policy from OpenMalaria simulations can be posed as a stochastic multi-armed bandit problem. For example this formulation has been used as an approach to develop models which may aid in the design of clinical trials, where actions should be made to balance exploitation (positive patient outcomes) and exploration (searching for actions which may lead to a clinical `breakthrough'). In our framing we wish to efficiently determine high performing policies for a simulated population of individuals over a 5 year intervention time frame. 
\begin{figure}[!t]
\centering
\tikzstyle{OMenvironment} = [rectangle, draw, fill=blue!20, 
    text width=6em, text centered, rounded corners, minimum height=4em]
\tikzstyle{line} = [draw, -latex']
\tikzstyle{Agent Model} = [draw, ellipse,fill=red!20, node distance=3cm,
    minimum height=2em]
    
\begin{tikzpicture}[node distance = 4cm, auto]

    \node [Agent Model] (Agent) {Agent Model};
    \node [OMenvironment, right of=Agent] (env) {OpenMalaria Simulation Environment};


    \path[line] (env) |-([shift={(3mm,-3mm)}]env.south west)-- node [shift = {(3mm,0mm)}]{$R_{\theta}(\bm{a}_i)$} ([shift={(3mm,-6.15mm)}]Agent.south)-|(Agent);
    
      \path[line] (Agent) |-([shift={(-3mm,7.25mm)}]Agent.north east)-- node [shift = {(-3mm,0mm)}]{$\bm{a}_i$}([shift={(-3mm,3mm)}]env.north)-|(env);

\end{tikzpicture}
\caption{Policies $\bm{a}_i$ are chosen by the Agent Model which receives rewards $R(\bm{a}_i)$}
\label{fig_flow}
\end{figure}
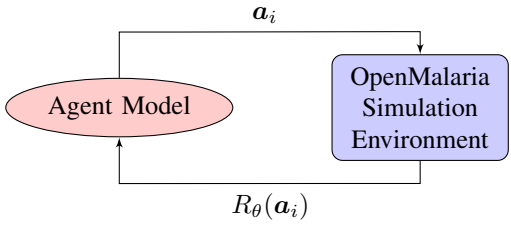

\subsection{State}
Due to the multi-armed bandit framing of the problem there is no state transition between OpenMalaria simulations. Instead we are trying to solve the problem of making one-shot policy recommendations for the simulation intervention period of 5 years. The state is therefore defined by the simulation's initial parameters $\theta$ and the policy $\bm{a}$ simulated.

\subsection{Action}
The main control methods used in Rachuonyo South district are: mass-distribution of long-lasting insecticide-treated nets (ITNs); Indoor Residual Spraying (IRS) with pyrethroids; and the prompt and effective treatment of malaria. This work will explore a policy space made up of the first two components (ITNs and IRS) which are direct intervention strategies, while prompt and effective treatment is described by the OpenMalaria simulation parameters and impacts the rewards detailed in the proceeding section. 
The domain of the first component is the deployment of nets, which defines the coverage of the population ($a_{\mathrm{ITN}} \in (0,1]$). 
The domain for the second component is the application of seasonal spraying, which defines the proportion of population coverage for this intervention ($a_{\mathrm{IRS}}\in (0,1]$). 
The spraying regimens of IRS are conducted through alternating the intervention between April/June each year \cite{Stuckey2014}. 
Since the policy decision is framed as how much of the simulated population should be covered by a particular intervention, the policy space $A$ is constructed through $\bm{a}_i \in A = \{a_{\mathrm{ITN}}, a_{\mathrm{IRS}}\}$.

For the studied scenario the simulation environment handles distribution of the interventions across the simulated population. The agent is not controlling the more complex actions of targeted interventions which have not been previously reported on, though it should be noted that while the action space is finite (there are a finite number of individuals in the simulation model), the size of this space will grow exponentially as more interventions or targeted interventions are added.
Simulation compute time also grows linearly with number of simulated individuals. As such, a complete exploration of the entire action space quickly becomes in-feasible as complexity grows toward any real-world equivalent simulation. What is presented here is a first approximation to a real-world scenario.

\subsection{Reward}
The reward associated with each policy $R_\theta(\bm{a}_i)$ is stochastic through the parameterisation of the simulation $\theta$, which generates a randomised distribution of parameters for the OpenMalaria simulation. The magnitude of the reward is determined through an economic cost-effectiveness analysis of the stochastic simulation output. The following sections give an overview of the calculations used, as specified in the health economics literature.

\subsubsection{DALYs}
Disability adjusted life years \cite{Murray1996}, are a measure defined by the total years of life lost (YLL) due to fatality linked with contraction of the disease, and number of years of life with disability (YLD) as a result of the disease. 
Upon termination, the OpenMalaria simulation produces outcomes for each individual in the population for the considered scenario.
If an individual experienced a malaria episode ($\tt{ME_k}$), the simulation results allow YLD to be quantified (See (\ref{eq:yld})).
Additionally if an individual contracted malaria and subsequently died ($\tt{D_z}$), either directly or indirectly from malaria, the simulation output allows YLL to be calculated (See (\ref{eq:ylli})). We also use a discount factor $\gamma = 0.97$ to discount the value of future years of life lost, and a life expectancy of 46.6 years \cite{network2004indepth}. The work of \cite{Briet2013} may be referred to for an explicit mapping of OpenMalaria simulation outputs to the calculation of DALYs.

\begin{align}
\YLD &=  \sum_{k=0}^K \tt{Duration}(ME_k)*\tt{Weight}(\tt{Age}(ME_k))\label{eq:yld}\\
\YLL_z &=  \mbox{max}(0,\tt{LifeExpectancy} - \tt{Age}(D_z)))\label{eq:ylli}\\
\YLL &=  \sum_{z=0}^Z \YLL_z \times \gamma^{\mathrm{YLL}_z} \label{eq:yll}\\
\DALY &= \YLL + \YLD \label{eq:daly}\
\end{align}

\subsubsection{Simulated Costs}
In this work we simulate two types of costs, the cost to treat and manage malaria episodes, and the cost to implement interventions which minimise malaria prevalence.
We call these healthcare system costs (HSC), and intervention costs (IC) respectively.

For each malaria episode that a patient seeks treatment, hospitals incur costs to treat the disease, to manage the patient's recovery process, and also to deal with the patient's death if that were to occur.
As such, the HSC can be broken down in terms of total in-hospital treatment costs (TTC), the total in-hospital recovery cost (TRC), and the cost for in-hospital mortality (See (\ref{ttc}-\ref{total})).
We use values from the literature \cite{Tediosi2009} to define the costs implemented in this work.
\begin{align}
	\TTC &=  \sum_{k=0}^K \tt{Cost}(\tt{Treatment}(\tt{InHospital}(ME_k))) \label{ttc}\\
	\TRC &=  \sum_{k=0}^K \tt{Cost}(\tt{Recovery}(\tt{InHospital}(ME_k)))\label{trc}\\
	\HSC &= \TTC + \TRC + \sum_{z=0}^Z \tt{Cost}(\tt{InHospital}(D_z))
	\label{total}
\end{align}

The cost of the intervention ($C_{\mathrm{int}}$) is the sum of the numbers of individuals covered by each intervention by the average cost of deploying the intervention to an individual. 
For the region in Kenya under consideration it costs 8.52USD per net and 0.73USD per person covered by spraying intervention. 
The average cost to seek hospital treatment per person is 0.60USD, and this value is associated with transportation and consumables. \cite{Stuckey2014}.

\subsubsection{Cost Effectiveness}
The agent models proposed will receive rewards based on the cost effectiveness of a policy, a metric often used by researchers evaluating the impact of a policy. 
We define cost effectiveness as the ratio of the relative cost to perform a policy intervention to the health impact realized from that policy intervention.
The health impact is defined as the DALYs averted ($\mathrm{DA}$), the difference between the DALYs realized with the application of the considered policy intervention and the DALYs realized when performing no intervention.
So the cost effectiveness will be quantified as the cost per DALY averted ($C_{\mathrm{DA}}$):
\begin{equation}
C_{\mathrm{DA}} =\dfrac{\HSC_{\mathrm{int}}-\HSC_{\mathrm{no int}} + C_{\mathrm{int}}}{\mathrm{DA}} 
\end{equation}

\subsection{Gaussian Process Regression}
The rewards received from the OpenMalaria simulation environment are stochastic as there is noise built in to the underlying models.
Despite stochastic simulation results, the rewards received from similar policies should be highly correlated.
If we consider that each simulated policy returns a stochastic scalar reward $R(\bm{a}^{1})...R(\bm{a}^{n})$ with mean $\mu(\bm{a}) = E[R(\bm{a})] $ and covariance $ k(\bm{a},\bm{a}')=E[(R(\bm{a})-\mu(\bm{a}))(R(\bm{a}')-\mu(\bm{a}'))]$, a gaussian process can be specified by these mean and covariance functions $GP(\mu(\bm{a}),k(\bm{a},\bm{a}'))$. 

Gaussian Process regression (GPR) is a supervised learning technique, in which the stochastic scalar rewards $R$, are used to train a Gaussian Process to infer with confidence bounds the performance of actions across the policy space \cite{Williams1996}.  
The learnt parameters describe the posterior distribution over $R(\bm{a})$:
\begin{equation}
\mu_{i+1}(\bm{a}) = \bm{k}_i(\bm{a})^{T}(\bm{K}_{i}+\sigma^2\bm{I})^{-1}R_i
\end{equation}
\begin{equation}
\sigma_{i+1}(\bm{a}) = k(\bm{a},\bm{a}') - \bm{k}_i(\bm{a})^{T}(\bm{K}_{i}+\sigma^2\bm{I})^{-1}\bm{k}_i(\bm{a})
\end{equation}

At each location $\bm{a} \in A$,  $\bm{k}_i(\bm{a})=[k(\bm{a}_i^k,\bm{a})]_{\bm{a}_i^k\in A_i}$ and $\bm{K}_i = [k(\bm{a},\bm{a}')]_{\bm{a},\bm{a}'\in A_i}$, here $\sigma^2$ is the likelihood variance of the GP posterior. For this specific problem we have used a 0-mean function $\mu_0 \equiv 0$ and a Matern-5/2 covariance function or kernel $k(\bm{a},\bm{a}')$, of length scale $=l$ and parameter $\nu = 5/2$.

\subsection{Agent Models}
Each agent performs sequential batch exploration, towards optimisation of an unknown stochastic reward function $R$. At each batch ($i$) we will choose $j=1,2,..,B$ policies $\bm{a}_i^j \in A$. Due to the computational expense of calculating $R(\bm{a}_i)$ and the size of the entire $A$, we wish to find solutions of maximal reward in as few batches $i$ as possible. The goal being to approximate $\bm{a}^* = \mbox{argmax}_{\bm{a} \in A}R(\bm{a})$ without prohibitively expensive computation for all possible policies, therefore using a subset $A_c \in A$ of the policy space.

\subsubsection{Upper/Lower Confidence Bound (GP-ULCB)}
\label{sec:ucb}
We introduce the Gaussian Process Upper/Lower Confidence Bound  (GP-ULCB) algorithm, inspired by Gaussian Process regression (GPR) and work on Upper Confidence Bound (UCB) solutions to the multi-armed bandit problem 
\cite{Auer2010}\cite{Auer2002}. This is a formulation which combines the natural confidence bounds of Gaussian Processes for stochastic multi-armed bandit problems, and variants have already been proposed in the form of GP-UCB \cite{Srinivas2009} and GP-UCB-PE \cite{Contal2013}. 
\begin{algorithm}[]
\SetAlgoLined
\KwResult{$\bm{a}_i$: batch $i$}
 Input: random discretised actions $\bm{a} \in A_c$\; GP priors $\mu_0 = 0$, $\sigma_0,l$\; $B= $batch size, $f_m= $mixing factor, $f_c= $masking factor\;\
 \For{i = 1,2,...}{
 reset: $\bm{a}_{\mathrm{upper}},\bm{a}_{\mathrm{lower}}, A_c$\;
 \For{j = 1,2,..,B.}{
 	\eIf{$j<B \times f_m$}{
 	$\bm{a}_i^j = \underset{a \in A_c}{\mbox{argmax}} \ \mu_{i-1}(\bm{a}) + \beta\sigma_{i-1}(\bm{a})$\\
 	mask: $\bm{a}_{\mathrm{upper}}$, $|\bm{a^j}-\bm{a}_{\mathrm{upper}}| < l \times f_c$\\
 	update: $\bm{a}_{\mathrm{upper}} \notin A_c$
 	}
  	{$\bm{a}_i^j = \underset{a \in A_c}{\mbox{argmin}} \ \mu_{i-1}(\bm{a}) - \beta\sigma_{i-1}(\bm{a})$\\
  	mask: $\bm{a}_{\mathrm{lower}}$, $|\bm{a^j}-\bm{a}_{\mathrm{lower}}| < l \times f_c$\\
 	update: $\bm{a}_{\mathrm{lower}} \notin A_c$
  	}
 }
 Return: $R_{\theta}(\bm{a}_i)$\\
 Update Posterior: mean $\mu_{i}(\bm{a})$, variance $\sigma_{i}(\bm{a})$
 }
 \caption{GP-ULCB}
\end{algorithm}

The algorithm is initialised with a random sample of a discrete policy space ($A_c$). Subsequent policies are chosen to further explore the policy space regressed by GPR on all preceding simulation runs.  The choice of using both \textit{upper} and \textit{lower} confidence bounds was made due to the stochastic nature of rewards.
Specifically, minima and maxima can readily occur in the policy space necessitating a search for both potentially optimal and bad strategies.
Also, by including the sampling of minima, the agent may present a risk adverse exploration of the policy space. 

\subsubsection{Genetic Algorithm}

A genetic algorithm (GA) was implemented to provide comparison of another `black box' optimisation technique for the exploration of the policy space.
The GA is a biologically inspired, population-based search technique \cite{Holland1992}, specifically a meta-heuristic inspired by the process of natural selection.
We use the reward generated for a policy as the measure of its fitness, and as OpenMalaria allows us to calculate a stochastic reward for each policy, there is noise in the fitness measure.

Given an evaluated population, in this case a set of policies and their stochastic rewards, the GA with then derive the next generation of the population.
In this work we begin this process with roulette wheel selection \cite{Goldberg1991} to select candidate policies.
This biases selection of good policies to pass their `genetic material' to the subsequent generation.
The probability of selection  $p^j$ of the $j^{th}$ policy in a generation (i.e. batch $i$), is defined in (\ref{weightedroulettewheel}), where $f^j$ is the fitness ($-R(\bm{a}_i)$ normalised $[0,1]$).
\begin{equation}
p^j = \frac{f^j}{\sum_{k=1}^{B} f^k}
\label{weightedroulettewheel}
\end{equation}

Each policy in the subsequent generation is derived from two policies selected via this approach.
Mimicking biological crossover of chromosomes, the two selected policies are mixed, and one of the resulting policies selected at random.
Finally, a random subset of the components of each derived policy is perturbed by adding noise.
This sequence of processes defines how policies from the current generation are used to derive the next generation.

\subsection{Batch Policy Gradient}
The final approach used was a modified policy gradient \cite{Sutton1999}, chosen as a method reported to handle continuous or very large action spaces in the case of this problem, while being robust to stochasticity. In this implementation $R_\theta(\bm{a})$ is approximated by a neural network, with new policies sampled through $\epsilon$-greedy exploration. The negated rewards $-R(\bm{a_i})$ were normalised $[0,1]$ from the batch results and the network trained to update it's weights ($\bm{w}$) associated with each policy $\bm{a} \in A_c$, using gradient descent on the negative log-loss of the batch normalised rewards.

During training the policy $\bm{a}^j$ (\ref{pgepsilon}) will be chosen with probability $\epsilon$:
\begin{equation}
\bm{a}^j = \underset{w \sim A_c}{\mbox{argmax}}(\bm{w})
\label{pgepsilon}
\end{equation}
While a random policy will be sampled from $A_c$ with probability $1-\epsilon$. Similarly to the GP-ULCB algorithm each $\bm{a}^j$ of batch $i$ is sampled sequentially such that $\bm{a}^{j-1} \notin A_c$.

\section{System Implementation and Deployment}
For this work we used OpenMalaria commit  a50730b. 
The simulation environment was run on a 4 node cluster of machines, each with 64 hyper threaded cores (2.20GHz Intel Xeon\textregistered  \ CPU E5-2660). On these processors, running one instance of an OpenMalaria simulation, for a representative human population size (100,000), returns results in the time-frame of days.
Running experiments in batches can thus take advantage of the natural parallelism of the deployment environment.
Parallelism was implemented using Python's \textit{multiprocessing} package.
This package supports spawning processes using an API, and was used to execute OpenMalaria simulations, passing the scenario as an argument. This results in a natural expression of the batch size $B$ for each agent model equal to the number of available processor cores.

\section{Results}
Published studies \cite{Stuckey2014} give a direct reference to human expert decisions made using OpenMalaria as a research tool, specifically to compare the cost-effectiveness of different interventions in Rachuonyo South District. Their findings stated that the current policy of  56\% $a_{\mathrm{ITN}}$, 70\% $a_{\mathrm{IRS}}$ was the most cost-effective with regards to $C_{\mathrm{DA}}$, while they also recommended that increasing this to 80\%  $a_{\mathrm{ITN}}$ and 90\% $a_{\mathrm{IRS}}$ (including a school-based screen and treat program) would have the greatest health impact for DALYs averted.
In this work we use the same stochastic parameterisation $\theta$ of OpenMalaria, but instead explore an automated answer to a less constrained problem for the decision maker: \textit{given the current intervention strategy, what policy decisions can be made for the next 5 years to improve cost-effectiveness}? 

Our results indicate that all three agent models extract the same top three performing emergent policies with respect to our primary evaluation metric, $C_{\mathrm{DA}}$ (See Table \ref{tab_results}):
\begin{itemize}
\item Maintain $a_{\mathrm{ITN}}$ and stop $a_{\mathrm{IRS}}$, 
\item Maintain $a_{\mathrm{ITN}}$ and reduce $a_{\mathrm{IRS}}$, 
\item Increase $a_{\mathrm{ITN}}$ and stop $a_{\mathrm{IRS}}$.
\end{itemize}

\begin{figure*}[ht]
\centering
\subfloat[Subfigure 1 list of figures text][Current and Expert Human recommended policies]{
\includegraphics[width=0.45\textwidth]{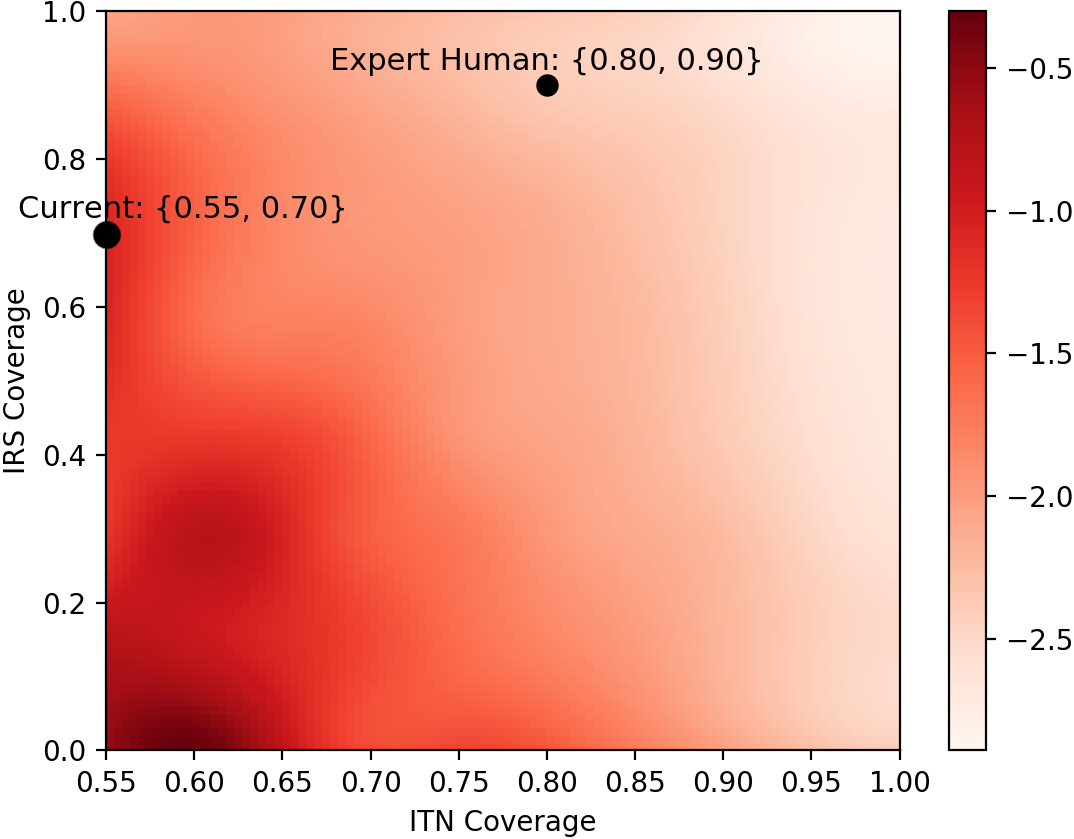}
\label{fig:subfig1}}
\subfloat[Subfigure 2 list of figures text][GP-ULCB ]{
\includegraphics[width=0.45\textwidth]{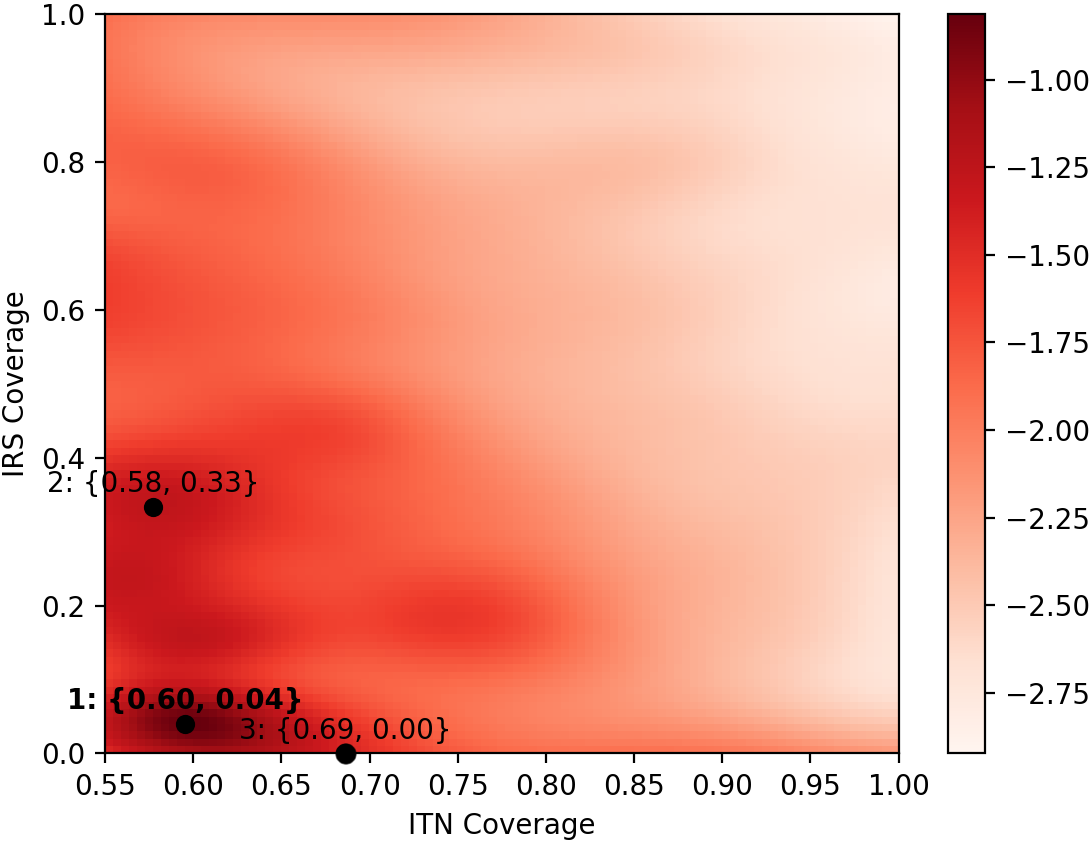}
\label{fig:subfig2}}
\qquad
\subfloat[Subfigure 3 list of figures text][Genetic Algorithm]{
\includegraphics[width=0.45\textwidth]{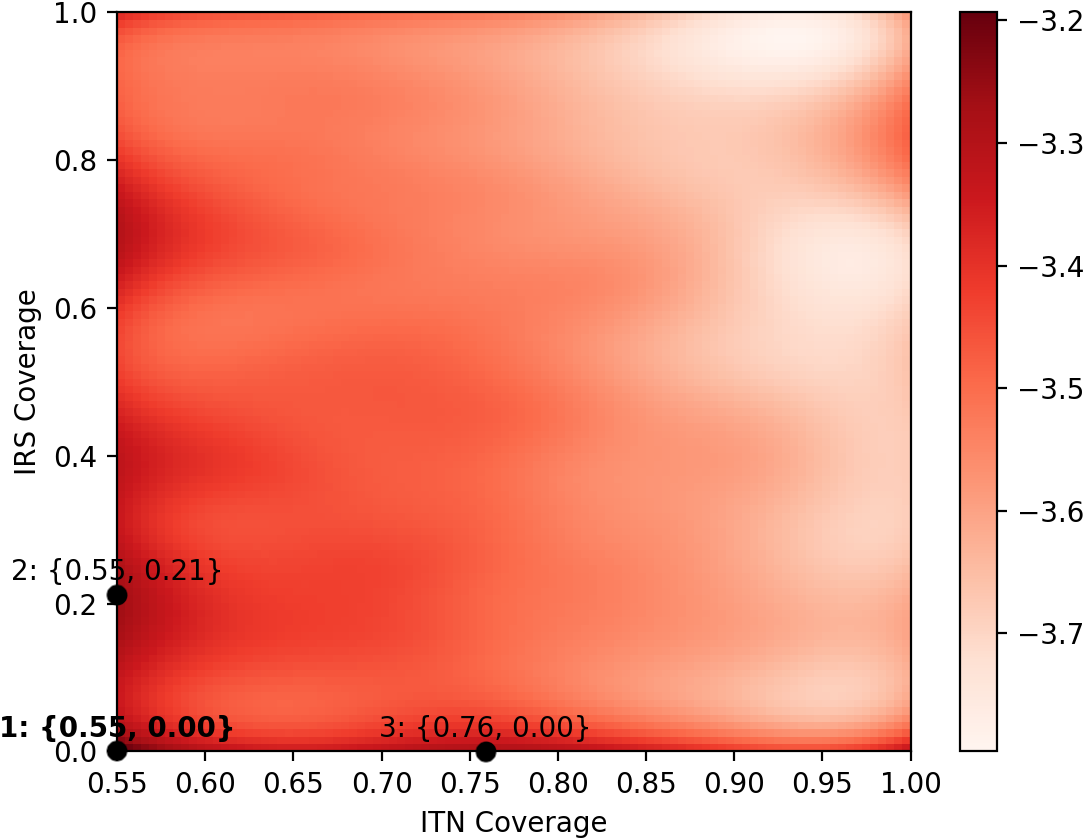}
\label{fig:subfig3}}
\subfloat[Subfigure 4 list of figures text][Batch Policy Gradient]{
\includegraphics[width=0.45\textwidth]{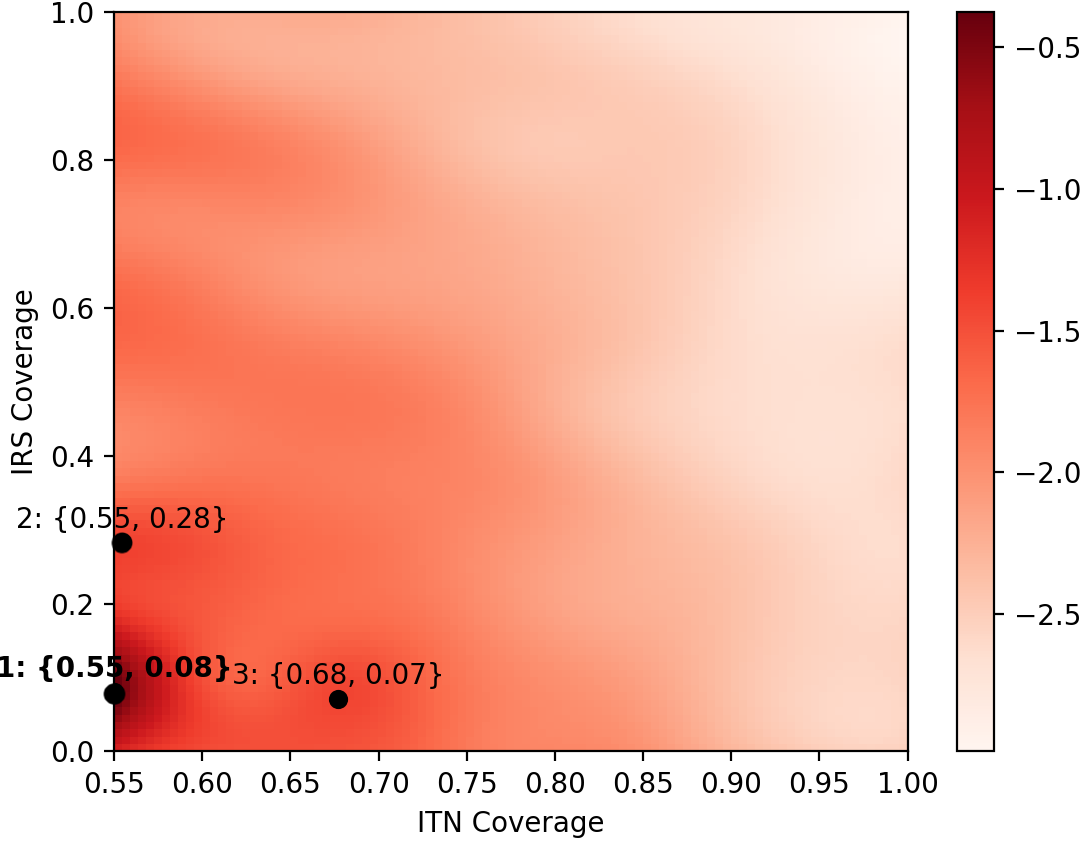}
\label{fig:subfig4}}
\caption{Visualisations of $-\log(R_\theta(\bm{a}))$ policy surfaces. Generated from each agent, after 8 iterations of a 64 batch size, running a total 512 simulations. All regressed with the same Gaussian Process parameters. The surfaces describe each agent's exploration of the available actions $a_{\mathrm{ITN}}$ and $a_{\mathrm{IRS}}$ for a human decision maker in Rachuonyo South.}
\label{fig:globfig}
\end{figure*}

These findings are extracted from the surface maxima of the posterior mean $\mu(\bm{a})$ 
through Gaussian Progress regression of rewards $R_\theta(\bm{a})$ collected by each respective agent.
Figure \ref{fig:globfig} illustrates these surfaces.

\begin{table*}\centering
\ra{1}
\begin{tabular}{@{}rrrrrcrrrcrrrr@{}}\toprule
\multicolumn{4}{c}{\textbf{GP-ULCB}\ref{sec:ucb}} & \phantom{}& \multicolumn{4}{c}{\textbf{Genetic Algorithm}} &
\phantom{} & \multicolumn{4}{c}{\textbf{Batch Policy Gradient}}\\
\cmidrule{1-4} \cmidrule{6-9} \cmidrule{11-14}
Policy& $C_{\mathrm{DA}}$ & $\mbox{DA}$ & $C_{\mathrm{int}}$ && Policy& $C_{\mathrm{DA}}$ & $\mbox{DA}$ & $C_{\mathrm{int}}$&&Policy& $C_{\mathrm{DA}}$ & $\mbox{DA}$ & $C_{\mathrm{int}}$\\ \midrule
$\{\bm{60},\bm{4}\}$ & 28.9 & 11800 & 514000 &&$\{\bm{55},\bm{0}\}$ & 25.5 & 8690  &458000&& $\{\bm{55},\bm{8}\}$ & 27.6 & 11600 & 477000 \\
$\{58,33\}$ & 30.1 &12300  &519000&& $\{55,21\}$ & 26.8 & 11600 & 477000&&$\{55,28\}$ & 30.2 & 11000 & 489000\\
$\{69,0\}$ & 30.9 & 13000 & 579000&& $\{76,0\}$ & 28.3 & 11700 & 632000 &&$\{68,7\}$ &30.3 & 13800 & 589000\\
\bottomrule
\end{tabular}
\caption{Top three policies with respect to $C_{\mathrm{DA}}$ evaluated by each agent model. Policy: $\{a_{\mathrm{ITN}} \%,a_{\mathrm{IRS}}\%\}$, $C_{\mathrm{DA}}$: Cost per DALY Averted USD, DA: DALYs Averted, $C_{\mathrm{int}}$: Intervention Costs USD.}
\label{tab_results}
\end{table*}

\section{Evaluation}
The methods shown give a comprehensive evaluation, exploring the cost-effectiveness for a policy of the two main malaria interventions. Such fundamental insight is often missing from empirical studies, which are grounded in determining how much of a single intervention may be implemented to maximise a particular performance metric.
Interestingly, at the surface our results challenge the current sentiment in the community - that policy makers in Sub-Saharan Africa should maximise ITN coverage \textit{before} looking into other intervention strategies.
One 2017 study even states that:
\begin{quote}
coverage of ITNs was
consistently the most cost-effective intervention across
a range of transmission settings and was found to
occur early in the cost-effectiveness scale-up pathway.
IRS, RTS, S and SMC entered the cost-effective
pathway once ITN coverage had been maximised. \cite{Winskill2017}
\end{quote} 

Instead, our results suggest that when the cost effectiveness of strategies is considered, after achieving a threshold of nets deployed for the environment, it may be more cost-effective to start spraying a small proportion of households (approx. 20-30\%), instead of continuing to scale the deployment of insecticide treated nets. 

If additional investment is available, then these resources should be allocated to scale up the coverage of bednets and further maximise health outcomes. 
With unlimited resources there are health benefits in the maximization of nets first, however as policy makers are facing tougher budget constraints, there are other factors that should influence the decisions they will make and any smaller investment can more effectively used to spray households.
Our findings are exciting, however our contributions are limited as the current approach is only preliminary.
We have not explored if interventions could be deployed at different times of year, or even if multiple policies could have been concurrently deployed in a population.
Further, the current simulation did not permit interventions to be targeted to specific subsets of the population (e.g. households with young children). 
Finally, this work is specific to one studied location in Western Kenya, and the generalisation of the insights from multiple agents gathering insights across expansive environments e.g. Sub-Saharan Africa is yet to be explored.

\section{Further AI Research}
The techniques presented have been selected and designed with the view to deployment on larger policy spaces as detailed in the Evaluation section. More compute time, expansive environments and policies are a requirement for the real-world human decision maker. While other data sources exist outside of the OpenMalaria simulation environment, notably malaria mapping projects which could serve as visual input, requiring function approximation through Deep Learning. There is also the existing opportunity to embed the agent model deeper into the simulation environment, passing control of simulation parameters in order to allow balancing of computational expense with efficient policy space exploration. This work is viewed as an emerging application for deploying further novel exploration techniques.

\bibliographystyle{IEEEtran}
\bibliography{iaai}

\end{document}